\documentclass{article}

\usepackage{graphicx}
\usepackage{multirow}
\usepackage{amsmath, amssymb, amsfonts}
\usepackage{amsthm}
\usepackage{mathrsfs}
\usepackage[title]{appendix}
\usepackage{xcolor}
\usepackage{textcomp}
\usepackage{manyfoot}
\usepackage{booktabs}
\usepackage{algorithm}
\usepackage{algorithmicx}
\usepackage{algpseudocode}
\usepackage{listings}
\usepackage{nicefrac}
\usepackage{microtype}
\usepackage{subcaption}
\usepackage{caption}
\usepackage{lmodern}
\usepackage{authblk} % for author affiliations
\usepackage{url}
\usepackage[numbers]{natbib}
\usepackage{hyperref}

\title{Joint Embedding Predictive Architecture for self-supervised pretraining on polymer molecular graphs}

\newcommand{\equalcontrib}{\textsuperscript{*}}
\newcommand{\emailg}{\textsuperscript{\dag}}

\author[1,2]{Francesco Piccoli\equalcontrib}
\author[1]{Gabriel Vogel\equalcontrib\emailg}
\author[1]{Jana M. Weber\textsuperscript{\dag}}

\affil[1]{Department of Intelligent Systems, Delft University of Technology, Delft, The Netherlands}
\affil[2]{Work done before joining Amazon}

\date{} % no date

\begin{document}

\maketitle

\begingroup
\renewcommand\thefootnote{}\footnotetext{* These authors contributed equally to this work.}
\footnotetext{\dag\ Emails: g.vogel@tudelft.nl, j.m.weber@tudelft.nl}
\endgroup

\begin{abstract}
Recent advances in machine learning (ML) have shown promise in accelerating the discovery of polymers with desired properties by aiding in tasks such as virtual screening via property prediction. However, progress in polymer ML is hampered by the scarcity of high-quality labeled datasets, which are necessary for training supervised ML models. In this work, we study the use of the very recent 'Joint Embedding Predictive Architecture' (JEPA), a type of architecture for self-supervised learning (SSL), on polymer molecular graphs to understand whether pretraining with the proposed SSL strategy improves downstream performance when labeled data is scarce. Our results indicate that JEPA-based self-supervised pretraining on polymer graphs enhances downstream performance, particularly when labeled data is very scarce, achieving improvements across all tested datasets.
\end{abstract}

\textbf{Keywords:} Synthetic Polymers, Self-supervised Graph ML, JEPA, Transfer Learning, Molecular Property Prediction

\section{Introduction}
% Synthetic polymers 
Synthetic polymers are one of the most widespread classes of materials, constituting an essential component of numerous commodities in everyday life and industry \cite{aldeghiGraphRepresentationMolecular2022, 
kuennethPolymerInformaticsMultitask2021, kuennethPolyBERTChemicalLanguage2023, zhaoReviewApplicationMolecular2023a}.
The large diversity of the chemical space of polymers provides an opportunity to design polymers whose properties match the demands of the application \cite{kuennethPolyBERTChemicalLanguage2023, sattariDatadrivenAlgorithmsInverse2021}. However, the high number of combinations of monomer compositions, higher-order topologies, and processing methods brings challenges in the effective navigation of this large search space \cite{kuennethPolyBERTChemicalLanguage2023, zhaoReviewApplicationMolecular2023a}. 

In recent years, machine learning (ML) has shown potential in the discovery of new materials, including polymers \cite{martinEmergingTrendsMachine2023, yanRiseMachineLearning2023}. ML methods are increasingly applied in polymer science, particularly in two key areas: virtual screening of predefined candidate structures to predict properties and inverse polymer design to generate novel structures with desired properties \cite{chenPolymerInformaticsCurrent2021, reiserGraphNeuralNetworks2022, sattariDatadrivenAlgorithmsInverse2021}.
%This study focuses on the former—virtual screening—which leverages ML models to identify promising polymer candidates based on their predicted properties.
However, the application of ML in polymer science is still in its infancy, primarily due to the scarcity of high-quality, large, publicly available labeled datasets. This limitation arises from the time- and cost-intensive procedure of generating labeled polymer data (via experimental synthesis and testing or accurate molecular simulations) \cite{patelFeaturizationStrategiesPolymer2022,martinEmergingTrendsMachine2023, zhaoReviewApplicationMolecular2023a}. To overcome the problem of limited labeled data, several common strategies have been applied in the polymer domain. Transfer learning involves pretraining models on polymer properties with abundant labeled data and fine-tuning them for properties with limited data~\cite{yamadaPredictingMaterialsProperties2019, zhangTransferringMolecularFoundation2023}. Multitask learning is an effective approach to train predictive models on multiple properties with varying levels of labeled data, leveraging interdependencies between these properties \cite{gurnaniPolymerInformaticsScale2023, kuennethPolymerInformaticsMultitask2021, queenPolymerGraphNeural2023}. Lastly, self-supervised learning (SSL) makes it possible to pretrain models on large volumes of unlabeled data through tasks defined directly on the input data. The learned representations can then be fine-tuned on smaller labeled datasets \cite{gao2024self, kuennethPolyBERTChemicalLanguage2023, xuTransPolymerTransformerbasedLanguage2023, zhangTransferringMolecularFoundation2023, zhou2025polycl}.

Among these strategies, SSL has been particularly transformative across different data structures such as images \cite{caron2021emerging, simCLR, he2022MAE, sslCVsurvey}, natural language \cite{devlin2018bert, radford2018gpt}, and graphs \cite{jinSelfsupervisedLearningGraphs2020, liuGraphSelfSupervisedLearning2023, xieSelfSupervisedLearningGraph2023}. In the molecular domain, graph-based SSL has shown considerable success with small molecules~\cite{sunMoCLDatadrivenMolecular2021, zhang2021motifSSLMol, rong2020SSLGraphTransformerMolecules}.

In the context of polymers, the focus has largely been on text-based SSL, learning representations through tasks derived from the textual pSMILES representation~\cite{kuennethPolyBERTChemicalLanguage2023, xuTransPolymerTransformerbasedLanguage2023, zhangTransferringMolecularFoundation2023, zhou2025polycl}, with limited exploration of graph-based SSL approaches. 
Polymer graphs beyond the polymer repeat unit graph, including weighted edges that describe monomer ensembles, their topology and their stochasticity, as proposed in \cite{aldeghiGraphRepresentationMolecular2022}, present unique structural characteristics that distinguish them from small molecular graphs, posing challenges for directly applying SSL techniques developed for small molecular graphs. A recent study proposed a self-supervised graph neural network for such polymer graphs~\cite{gao2024self}. The authors employ two SSL tasks: one at the node/edge level, masking nodes and edges and learning to predict them, and the other at the graph level, predicting a pseudolabel corresponding to the molecular weight of the polymer, derived from the monomers' weights. They test both tasks separately and together, and they discover that pretraining via both tasks proves to be the most effective. This result aligns with findings in the literature \cite{hu2019strategies} that SSL on graphs works better when using both node-level and graph-level tasks together.

In this work, we study a new architecture family, called Joint Embedding Predictive Architecture (JEPA) \cite{lecunPathAutonomousMachine}, which was developed for self-supervised representation learning of images. Unlike traditional graph-based SSL methods, such as node or edge masking, which focus on reconstructing masked features directly in the input graph space, JEPAs operate in an embedding space. Predicting in the embedding space facilitates the learning of semantically-rich representations, avoiding the need to predict and reconstruct every (potentially noisy and hard to predict) detail of the input space, that in high-dimensional domains often leads to overfitting \cite{lecunPathAutonomousMachine, skenderiGraphlevelRepresentationLearning2023}. The way JEPAs learn, is by predicting the embedding of a "target view" of the graph based on the embedding of a "context view", typically by employing two encoders. 
We apply this architecture for self-supervised pretraining on stochastic polymer graphs to improve the accuracy in downstream tasks (e.g. property prediction) in label-scarce data scenarios.

We first use the proposed method for pretraining on a larger unlabeled corpus of data, and then finetune the model on available labeled data in a supervised fashion. While some aspects of our analysis apply broadly to JEPAs across various types of graphs (i.e. in different domains) and extend the study of JEPAs for graphs initiated in \cite{skenderiGraphlevelRepresentationLearning2023}, other results and experiments are specific to JEPAs in the molecular graph domain, specifically for stochastic polymer graphs.\\
%The results, discussed in Section \ref{results}, show that our pretraining strategy improves downstream performance, especially in data-scarce scenarios (less than 1000 datapoints), showing significant improvements also when finetuning on a dataset spanning a polymer chemical space different from the one employed for pretraining.

\section{Methods}
\label{methods}

\subsection{Polymer representation and datasets}\label{sec:data}
\begin{figure}[!htb]
    \centering
    \includegraphics[width=0.8\textwidth]{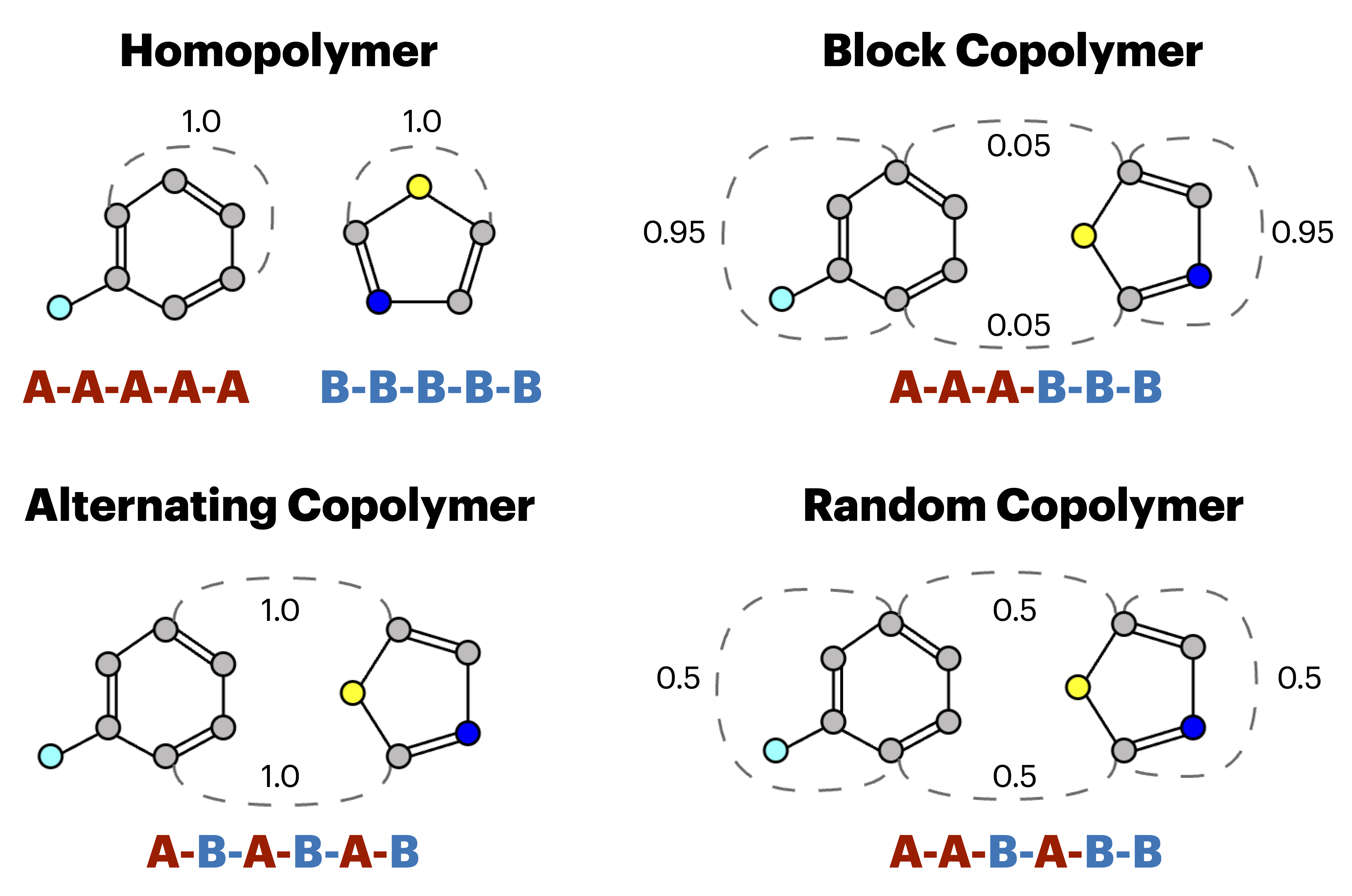}
    \caption{Polymer graph representation as introduced in~\cite{aldeghiGraphRepresentationMolecular2022}. The representation uses stochastic edges (dashed) reflecting the connection probabilities between monomers, i.e. reflecting the stochastic nature and chain architecture. }
    \label{fig:polymer_graph_simple}
\end{figure}

We represent polymers as stochastic graphs, as proposed in \cite{aldeghiGraphRepresentationMolecular2022}. The graph representation, visualized in Figure~\ref{fig:polymer_graph_simple}, connects monomer graphs through
weighted edges indicating the probabilities of connections, representing the polymer chain architecture. 

\begin{figure}[!htb]
    \centering
    \begin{subfigure}[b]{\textwidth}
        \centering
        \includegraphics[width=\textwidth]{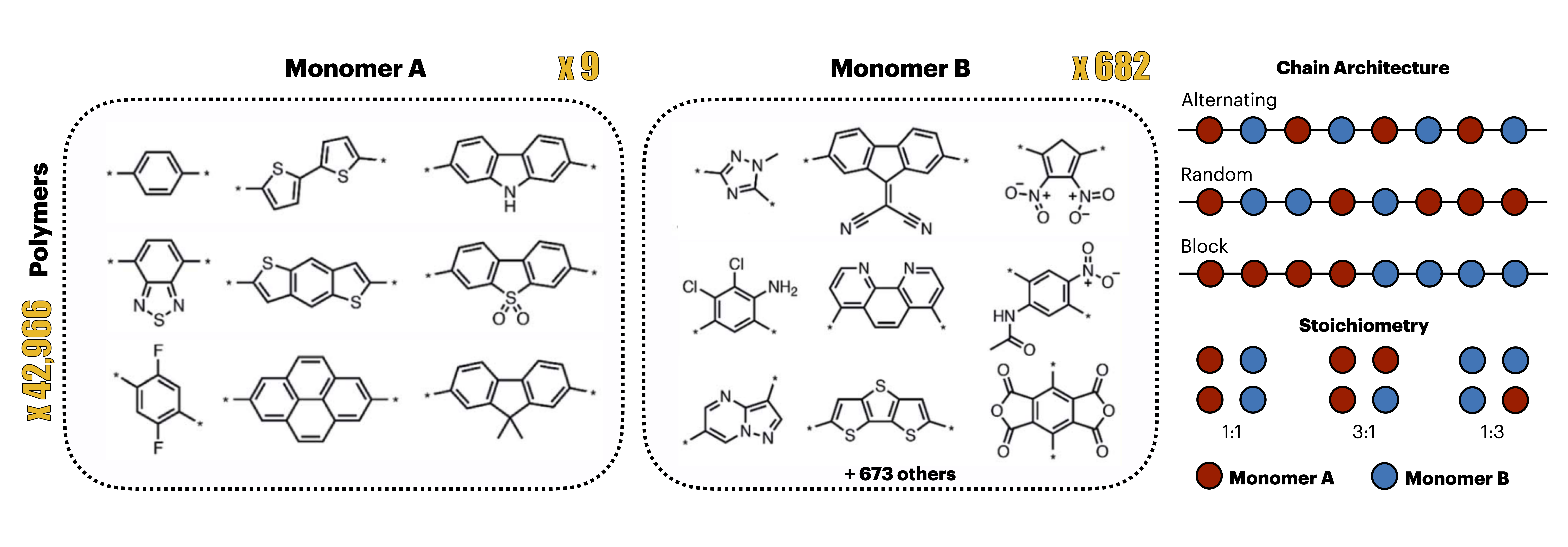}
        \caption{}
        \label{fig:aldeghi_dataset}
    \end{subfigure}
    
    \vspace{0.5em}

    \begin{subfigure}[b]{\textwidth}
        \centering
        \includegraphics[width=\textwidth]{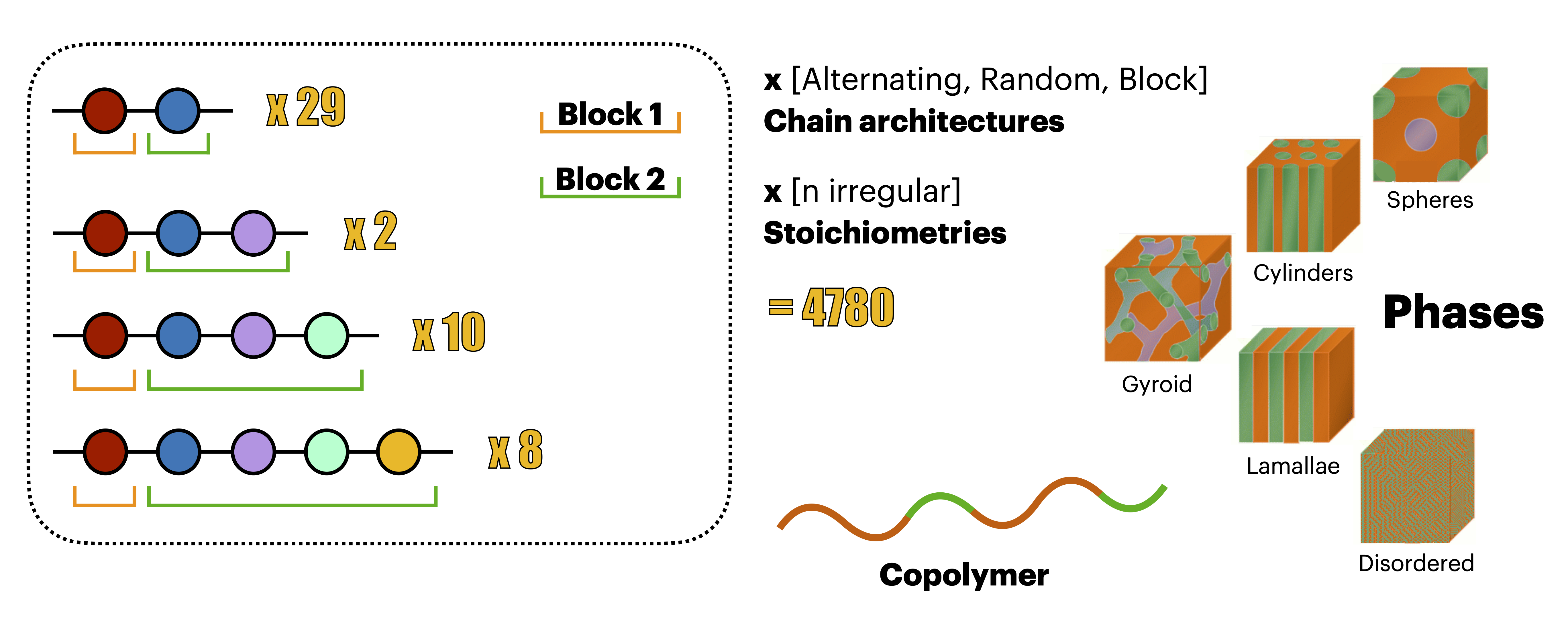}
        \caption{}
        \label{fig:diblock_dataset}
    \end{subfigure}
    
    \caption{(a) Conjugated copolymer photocatalyst dataset~\cite{aldeghiGraphRepresentationMolecular2022} with different stoichiometries and chain architectures. (b) Diblock copolymer dataset compiled by Arora et al.~\cite{arora2021dataset}, reporting experimentally observed phase behavior (e.g., lamellae, cylinders, gyroid) for 50 diblock copolymers across varying stoichiometries and chain architectures.}
    \label{fig:representation_dataset}
\end{figure}

We use the dataset of conjugated copolymer photocatalysts for hydrogen production\cite{aldeghiGraphRepresentationMolecular2022}, that is built upon the polymer space defined in \cite{bai2019accelerated}. As shown in Figure~\ref{fig:aldeghi_dataset}, the dataset contains 42,966 polymers, composed of nine A monomers and 862 B monomers. The polymers are classified according to three distinct chain architectures: alternating, random, and block. Additionally, they are further distinguished by three stoichiometry ratios: 1:1, 1:3, and 3:1. The authors of this dataset further published two calculated properties, the electron affinity (EA) and ionization potential (IP), for each of the copolymers~\cite{aldeghiGraphRepresentationMolecular2022}. We use this dataset for pretraining and keep aside a part of the dataset for finetuning. \\
We test our approach on two distinct downstream tasks, both starting from a model pretrained on the aforementioned dataset. Firstly, we finetune on data from the same dataset (but different split). Secondly, we finetune on a different downstream task using a dataset of Diblock copolymers \cite{arora2021dataset}. The dataset provides labels of the phase behavior (lamellae, hexagonal-packed cylinders, body-centred cubic spheres, a cubic gyroid, or disordered) of 49 diblock copolymers across various relative volume fractions, totaling 4780 labeled polymer samples (see Figure~\ref{fig:diblock_dataset}).

\subsection{Model} The idea of JEPAs for graphs is to first partition the graphs into patches (i.e. subgraphs) and define larger context subgraph $x$ and smaller target subgraphs $y$. Secondly, the goal is to predict (reconstruct) the embedding $\mathbf{s}_y$ of a target subgraph, from the embedding $\mathbf{s}_x$ of the context subgraph, operating in the embedding space (see Figure~\ref{fig:model}).  

\subsubsection{Subgraphing}
Before training the JEPA architecture, the polymer graphs need to be partitioned into a context subgraph and one or more target subgraphs. To achieve this, we first partition a polymer graph $G$ into a set of subgraphs $\{G_1, G_2, ..., G_n\}$, respecting the requirements outlined in Appendix~\ref{app:subgraph_reqs}. For this task, we explored three different subgraphing algorithms, i.e. random walk subgraphing, motif-based subgraphing and the METIS algorithm~\cite{METIS}. Context and target subgraphs are then selected (or created by combining subgraphs) from the set $\{G_1, G_2, ..., G_n\}$. In Section~\ref{sec:results_subgraphing}, we show the effect of different subgraphing algorithms, varying context and target subgraph size in percent of the whole polymer graph, and the number of targets on the model's performance. Each subgraphing algorithm has its own advantages and disadvantages:  
\begin{itemize}
    \item Motif-based subgraphing is a domain-specific approach to generate chemically meaningful subgraphs, such as functional groups or molecular subunits. We build on the BRICS (Breaking of Retrosynthetically Interesting Chemical Substructures) algorithm~\cite{degenArtCompilingUsing2008}, specifically using the implementation in~\cite{rBRICS}. The motif-based subgraphing method ensures meaningful subgraphs but potentially limits the variability of subgraphs due to its deterministic nature.\\
    \item METIS~\cite{METIS} is a popular subgraphing algorithm, using a clustering-based method, that partitions graphs into clusters while minimizing edge cuts and maximizing within-cluster links. Its widespread use is due to its low computational cost and the quality of the produced subgraphs. Despite being computationally efficient, METIS-based subgraphs lack chemical meaning compared to motif-based subgraphing.\\
    \item Random-walk subgraphing uses a stochastic approach to generate diverse subgraphs. It ensures greater flexibility and control over subgraph sizes while producing varying subgraphs at each training iteration. This method aligns well with the requirements for JEPA, particularly the need for dynamic changes to prevent overfitting.\\
\end{itemize}

\subsubsection{JEPA architecture for polymer graphs}
\begin{figure}[!htb]
    \centering
    \includegraphics[width=\textwidth]{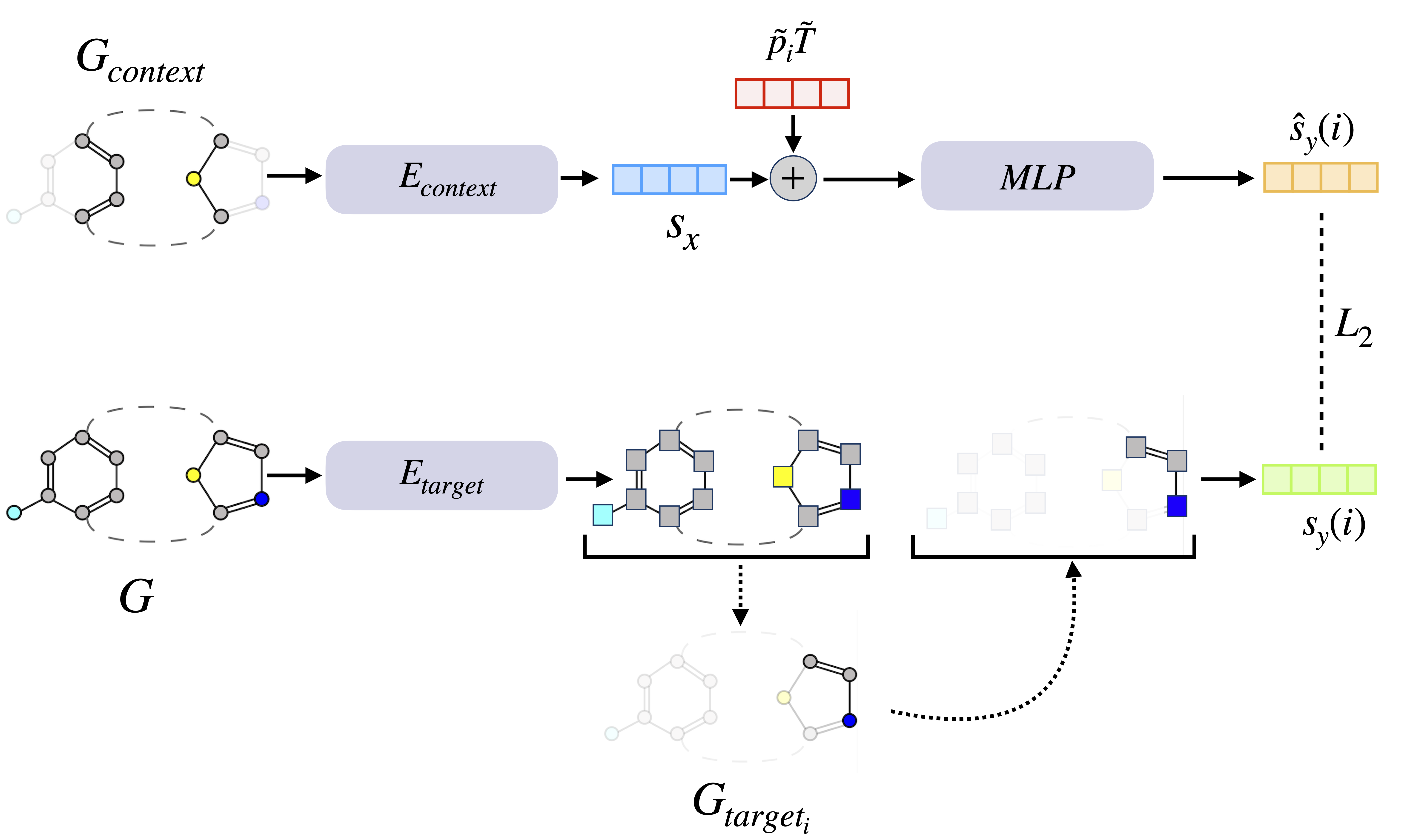}
    \caption{The polymer-JEPA model architecture. The model learns to reconstruct the embedding $\mathbf{s}_y(i)$ of a target subgraph $G_{target_i}$ from the embedded context graph $\mathbf{s}_x$, using a predictor network that is conditioned the positional encoding $\mathbf{\tilde{p}}_i \tilde{\mathbf{T}}$ to facilitate prediction. The loss is measured as the $L_2$ loss between the embedding $\mathbf{s}_y(i)$ and the predicted embedding $\mathbf{\hat{s}}_y(i)$.}
    \label{fig:model}
\end{figure}

The model architecture, depicted in Figure \ref{fig:model}, consists of two GNNs, the context and target encoders. As GNN architecture we utilize a variant of the weighted directed message passing neural network (wD-MPNN) as introduced by the authors of \cite{aldeghiGraphRepresentationMolecular2022}, however, utilizing node-centred instead of the original edge-centred message passing. In Appendix~\ref{app:node_vs_edge_WDMPNN}, we show that this change does not change the performance. The context encoder takes as input the context subgraph, which captures relevant local information. The target encoder, on the other hand, takes as input the entire polymer graph, allowing for an effective exchange of global information during the node embeddings generation via message passing, enabling effective contextualization of target embeddings. Given the modest size of the polymer graphs (20-30 nodes), the model can achieve this contextualization efficiently without relying on self-attention mechanisms. To obtain the context subgraph embedding \( \mathbf{s}_x \), the context encoder pools all the obtained node embeddings. To obtain the target subgraph embedding \( \mathbf{s}_y \), the target encoder pools the node embeddings learned from the target encoder for the nodes that belong to the target subgraph. Importantly, the subgraphs used to create the context subgraph cannot be selected as targets, which preserves the integrity of the prediction task. \\ 
The next step focuses on the prediction of the target subgraph embeddings through a multi-layer perceptron (MLP) from the context embedding and positional information of the target. We employed positional encoding (PE) via random-walk structural encoding (RWSE)~\cite{dwivedi2021graph, GraphViTMLPMixer2023} at two levels: at the node level and at the subgraph (patch) level. The node-level PE allows us to maintain node positional information when working with subgraphs. The subgraph (patch)-level positional encoding contains information about the connectivity of two subgraphs, here the context and target subgraph. 
Given the output of the context encoder, \(\mathbf{s}_x\), we wish to predict the \(m\) target subgraphs representations \(\mathbf{s}_y(1), \dots, \mathbf{s}_y(m)\). To that end, for a given target subgraph embedding \(\mathbf{s}_y(i)\), the predictor \(h_{\phi}\) takes as input \(\mathbf{s}_x\) summed with the linearly transformed target subgraph positional token \(\mathbf{\tilde{p}}_i\):
\begin{equation}
\mathbf{\hat{s}}_y(i) = h_{\phi}\left(\mathbf{s}_x + 
\mathbf{\tilde{p}}_i \tilde{\mathbf{T}}\right)
\end{equation}
with \(\tilde{\mathbf{T}} \in \mathbb{R}^{\tilde{k} \times d} \), where $\tilde{k}$ is the dimension of the positional encoding token and $d$ is the embedding dimension. The predictor outputs the predicted target embedding \(\mathbf{\hat{s}}_y(i)\). Since we wish to make predictions for \(m\) target blocks, we apply our predictor \(m\) times, obtaining predictions \(\mathbf{\hat{s}}_y(1), \dots, \mathbf{\hat{s}_y}(m)\). In practice, the predictor \(h_{\phi}\) is implemented via a MLP. For each data point, the loss is the average \( L_2 \) distance (Mean Square Error (MSE)) between the $m$ predicted target subgraph representations and the $m$ true target subgraph representations.

\subsection{Training procedure}\label{sec:training_procedure}
We split the full conjugated copolymer dataset~\cite{aldeghiGraphRepresentationMolecular2022} in three parts: 40\% for pretraining, 40\% for the different finetuning scenarios, and the remaining 20\% for testing the property prediction performance. 
More specifically, the pretraining phase entails training the JEPA architecture on 40\% (17186 entries) of the conjugated copolymer dataset~\cite{aldeghiGraphRepresentationMolecular2022}. After pretraining, only the trained target encoder is utilized in the finetuning step to obtain the polymer graph embedding.
For the downstream task (e.g. polymer property prediction), an MLP is employed on top of the polymer graph embedding obtained from the target encoder. Finetuning is done end-to-end, wherein not only the MLP but also the target encoder weights are updated during the optimization process. This allows the polymer graph embeddings to also finetune to the specific downstream task. As mentioned in Section~\ref{sec:data}, we perform the finetuning on the two different datasets to investigate the difference between using the same chemical space for pretraining and finetuning in contrast to using two different polymer chemical spaces. The results for this study are presented in Section~\ref{sec:results_downstream} and Section~\ref{sec:results_downstream_diblock}.

\subsection{Additional pseudolabel objective}\label{sec:PL_method} 
\begin{figure}[!htb]
    \centering
    \includegraphics[width=0.49\textwidth]{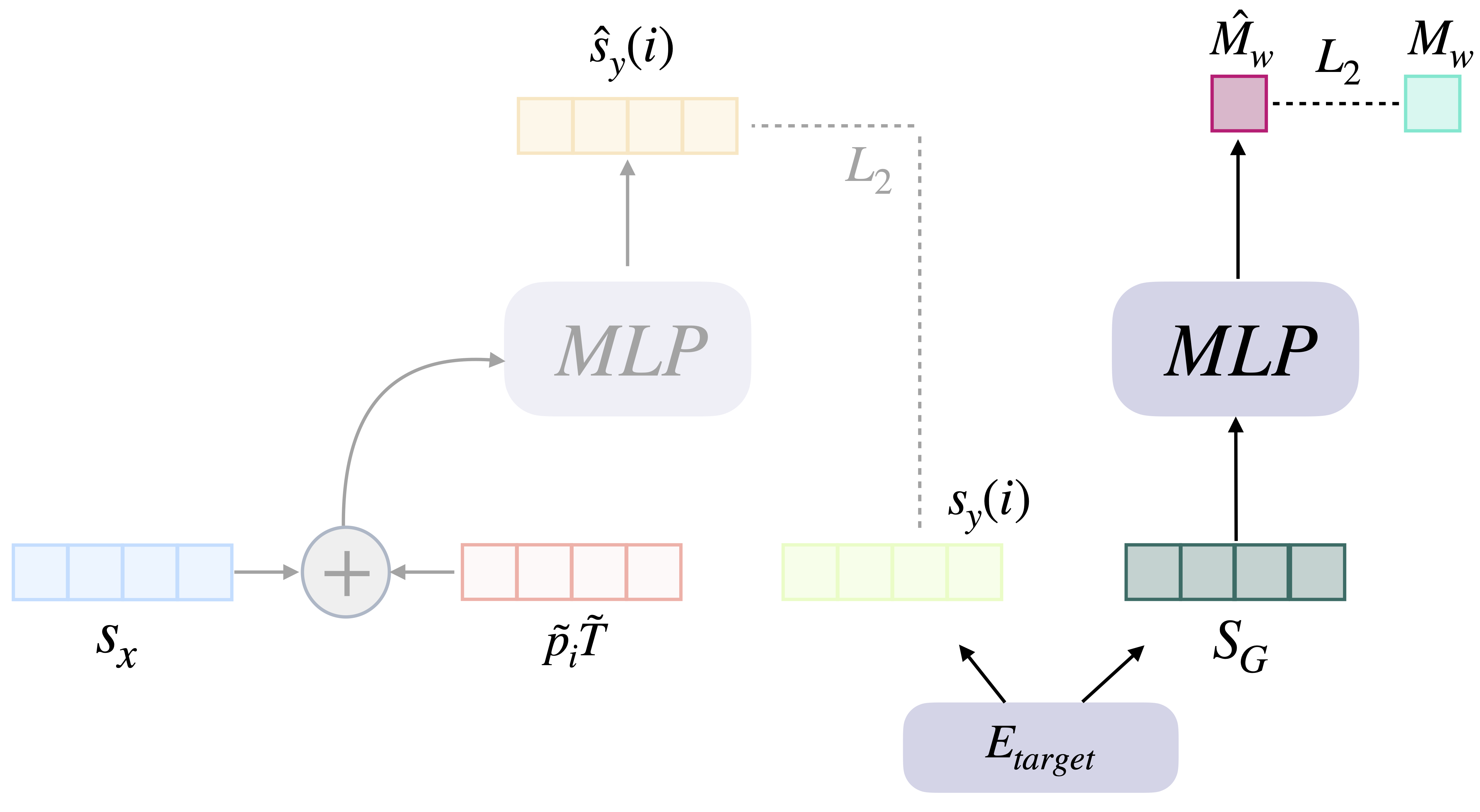}
    \caption{Pretraining with the additional pseudolabel objective. In the left, faded out, part, the embeddings from the target encoder $E_{target}$ are used to calculate the $L_2$ loss with the predicted target embeddings in the JEPA architecture. In the right part the polymer graph embedding is used as input for the MLP that predicts the pseudolabel (polymer molecular weight $M_w$). }
    \label{fig:pseudolabel}
\end{figure}

As mentioned in the introduction, a first study on self-supervised learning on the stochastic polymer graph representation~\cite{aldeghiGraphRepresentationMolecular2022} used both node and edge masking and prediction of a pseudolabel as self-supervised tasks. The study found that transferring not only the pretrained weights of the wD-MPNN encoder (equivalent to our target encoder), but also the weights of the pseudolabel predictor (an MLP used to predict molecular weight during pretraining) improved downstream performance. These pseudolabel predictor weights of the MLP were reused for downstream tasks, such as predicting electron affinity (EA) in the conjugated copolymer dataset~\cite{aldeghiGraphRepresentationMolecular2022}. Motivated by these results, we decided to adopt the pseudolabel objective also with our architecture. We do so by predicting the molecular weight from the polymer fingerprint learned through the target encoder, as visible in Figure \ref{fig:pseudolabel}. Unlike the sequential training approach in~\cite{gao2024self}, which first trains on node-level tasks and then on the pseudolabel task, we jointly train the encoder with both the JEPA objective and the pseudolabel prediction task. The molecular weight is defined as the weighted (monomer stoichiometry) sum of the molecular weights of the monomers.

\section{Results and Discussion}
\label{results}
In this section, we analyze the performance of our JEPA-based self-supervised pretraining strategy on polymer graphs. Section~\ref{sec:results_downstream} evaluates the model’s performance on downstream tasks under varying labeled data availability scenarios. 
As indicated in Section~\ref{sec:data}, we test the effectiveness of the self-supervised pretraining for predicting the electron affinity (EA) (regression) of the conjugated copolymer dataset (Section~\ref{sec:results_downstream}) and the phase behavior (classification) of diblock copolymers (Section~\ref{sec:results_downstream_diblock}). The latter task is based on a different polymer chemical space than the pretraining dataset, and thus evaluates the model's transfer learning capability.
Further, Section~\ref{sec:comparison_results_gao} compares our approach with an alternative self-supervised learning (SSL) method for polymer graphs, while Section~\ref{sec:rf_comparison} shows the performance against a simple baseline random forest model. Finally, we present an ablation study of subgraphing hyperparameters in Section~\ref{sec:results_subgraphing}, which provides general insights for JEPA-based graph machine learning.

\subsection{Downstream performance on data set of conjugated copolymers}\label{sec:results_downstream} 

\begin{figure}[!htb]
    \centering
    \includegraphics[width=0.6\textwidth]{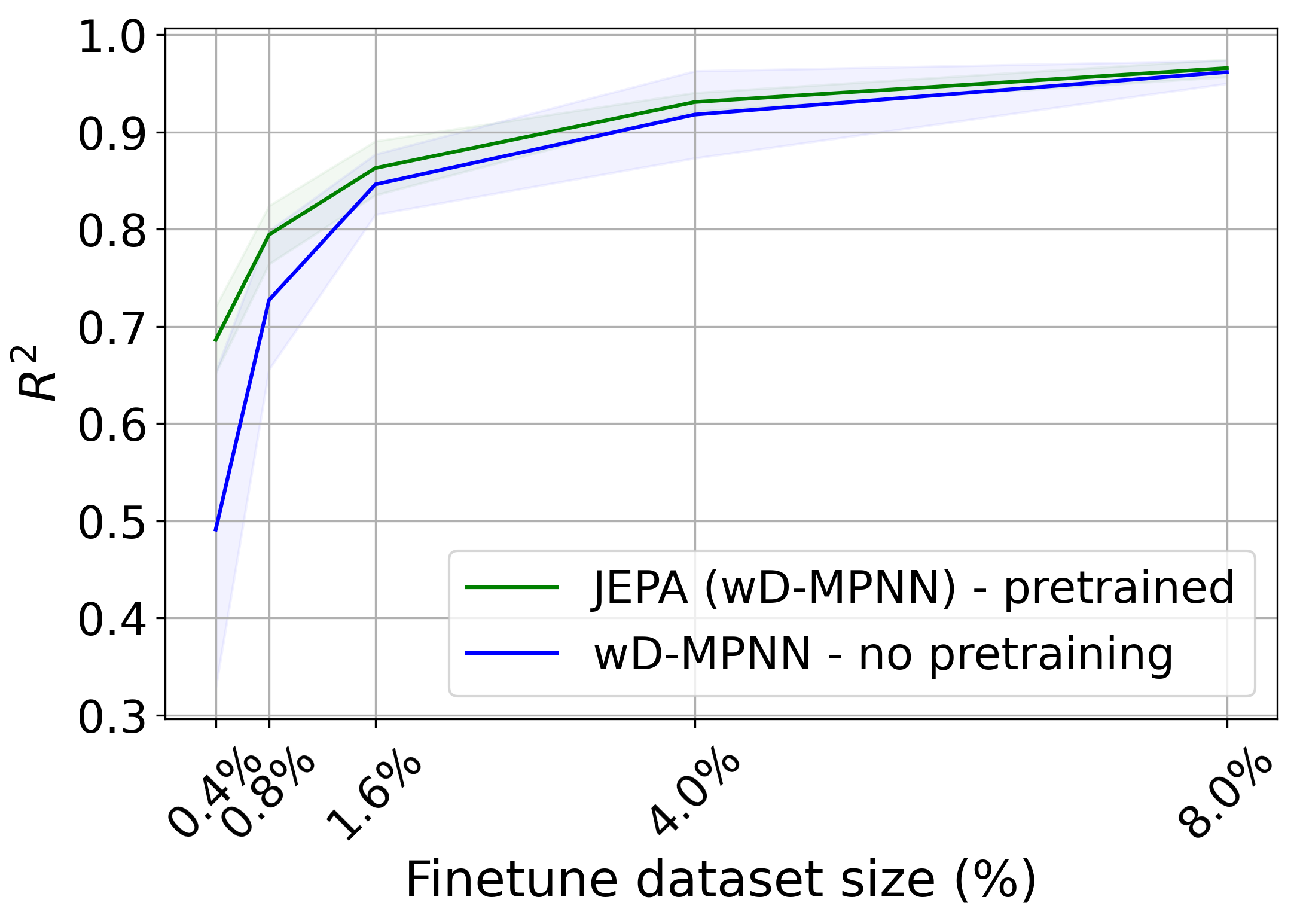}
    \caption{Effectiveness of our pretraining strategy for different finetune dataset sizes. The performance is evaluated on predicting the electron affinity of test data from the conjugated copolymer dataset~\cite{aldeghiGraphRepresentationMolecular2022} with the performance measured as $R^2$.}\label{fig:pretrain_nopretrain_aldeghi_EA}
\end{figure}

For the task of predicting the electron affinity (EA) of conjugated copolymers, our pretraining approach improves the performance especially in low labeled data scenarios as shown in~Figure~\ref{fig:pretrain_nopretrain_aldeghi_EA}. We test from a data scenario of 0.4\% (192 polymers) labeled data points to a scenario of 24\% (10,311 polymers) labeled data points. In Figure~\ref{fig:pretrain_nopretrain_aldeghi_EA} we report results only up to the 8\% scenario to better visualize the impact in low data regimes.
The pretrained model especially demonstrates performance improvements in scenarios up to a data size of 4\% (1728 polymers) labeled data points. However, beyond this threshold, the benefits of pretraining plateau. 
This suggest that the available labeled data is sufficient for supervised learning, rendering the transferred pretraining knowledge redundant. In practice, a small change in the R2 value (e.g. $\pm0.01$) does not significantly impact molecular design task, however, in the low labeled data scenarios (i.e. 0.4\% and 0.8\%), pretraining leads to a significant improvement of the property prediction performance. 

\subsection{Self-supervised transfer learning performance on Diblock copolymer dataset}\label{sec:results_downstream_diblock}
\begin{figure}[!htb]
    \centering
    \includegraphics[width=0.6\textwidth]{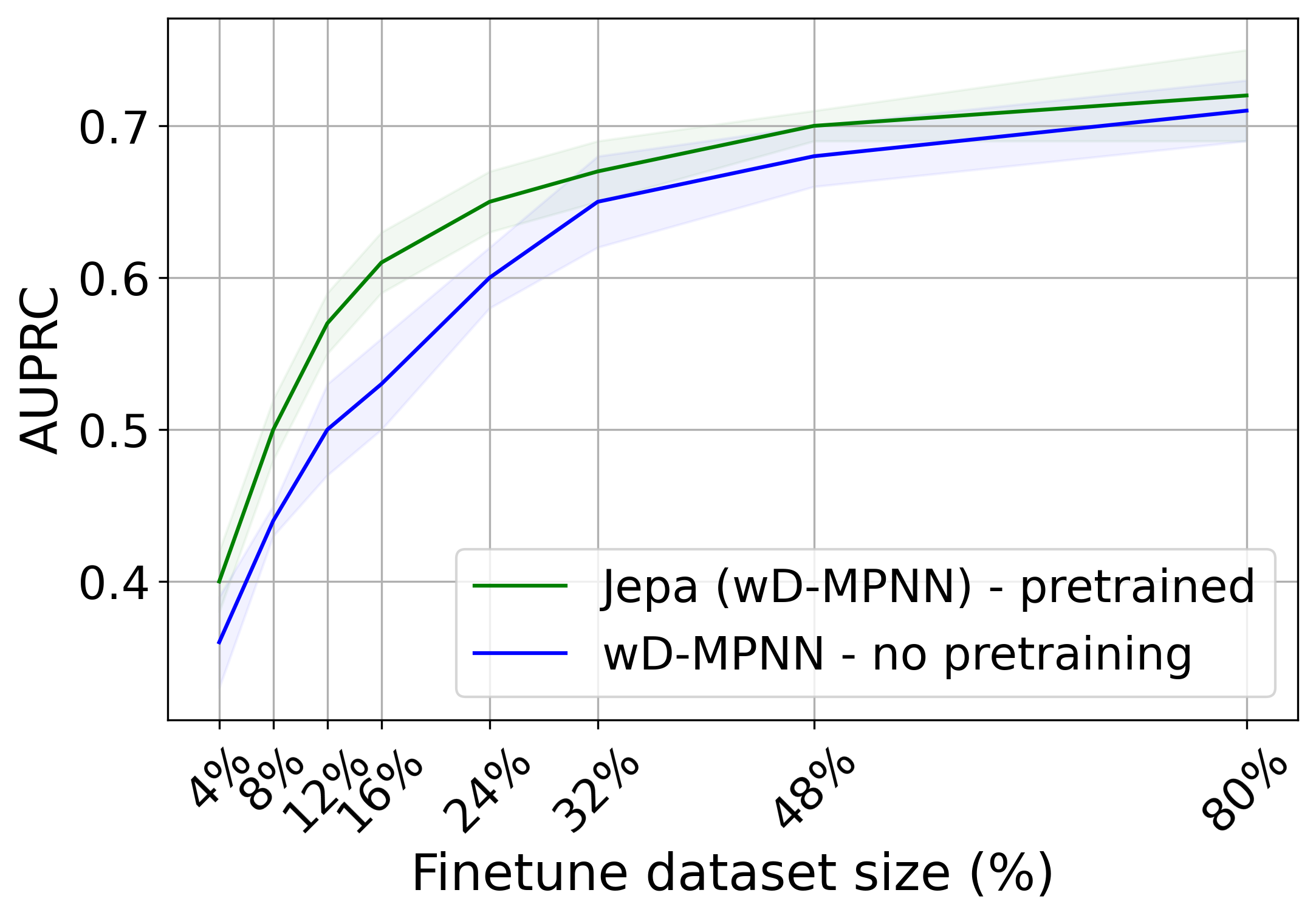}
    \caption{Effectiveness of our pretraining strategy for transfer learning and different finetune dataset sizes. Self-supervised pretraining is performed on data from the conjugated copolymer dataset~\cite{aldeghiGraphRepresentationMolecular2022} and finetuning is performed on test data from the diblock copolymer dataset~\cite{arora2021dataset}. The classification performance for predicting the phase behavior is measured as AUPRC (area under the Precision-Recall Curve).}    \label{fig:pretrain_nopretrain_diblock}
\end{figure}

For the task of predicting the diblock copolymer phase behavior, pretraining on the dataset of conjugated copolymers consistently improves the classification performance (between around 0.02 and 0.1 in AUPRC), even in higher labeled data scenarios, as illustrated in Figure~\ref{fig:pretrain_nopretrain_diblock}. We test from a training data scenario of 191 data points (4\%) to 3,824 data points (80\%). The latter scenario corresponds to finetuning on the full dataset, retaining 20\% for testing, as done in \cite{aldeghiGraphRepresentationMolecular2022}. %The results consistently demonstrate an improvement in the classification performance (between around 0.02 and 0.1 in AUPRC) in the pretrained scenario, even for the more data-rich scenarios. %This underscores the effectiveness of the proposed pretraining strategy and its capability to generalize outside of the training distribution. 
Since the pretraining dataset represents a different polymer chemical space compared to the finetuning dataset, we conclude that the knowledge acquired during pretraining is not overfitting or memorizing the training distribution (i.e. chemical space) but rather learning general chemical knowledge about polymers. This indicates the potential opportunity to utilise this strategy more broadly across different polymer data sets. 

We see particular promise in self-supervised transfer learning scenarios, where a model pretrained on a large, unlabeled polymer space is fine-tuned on a smaller, labeled dataset from another polymer space. We expect this to be especially advantageous for small datasets with complex structure–property relationships.

\subsection{Comparison to input space SSL}\label{sec:comparison_results_gao}
We compare the effectiveness of our pretraining strategy to the SSL pretraining method by Gao et al~\cite{gao2024self}, which uses node and edge masking in the input spaces as described in the introduction. Due to the small finetune dataset size, we run both methods with a 5-fold cross validation and three repetitions with different splits to ensure a robust comparison. The size of the wD-MPNN is set to three layers a hidden dimension of 300 for both methods. 

\begin{figure}[!htb]
    \centering
    \includegraphics[width=0.7\textwidth]{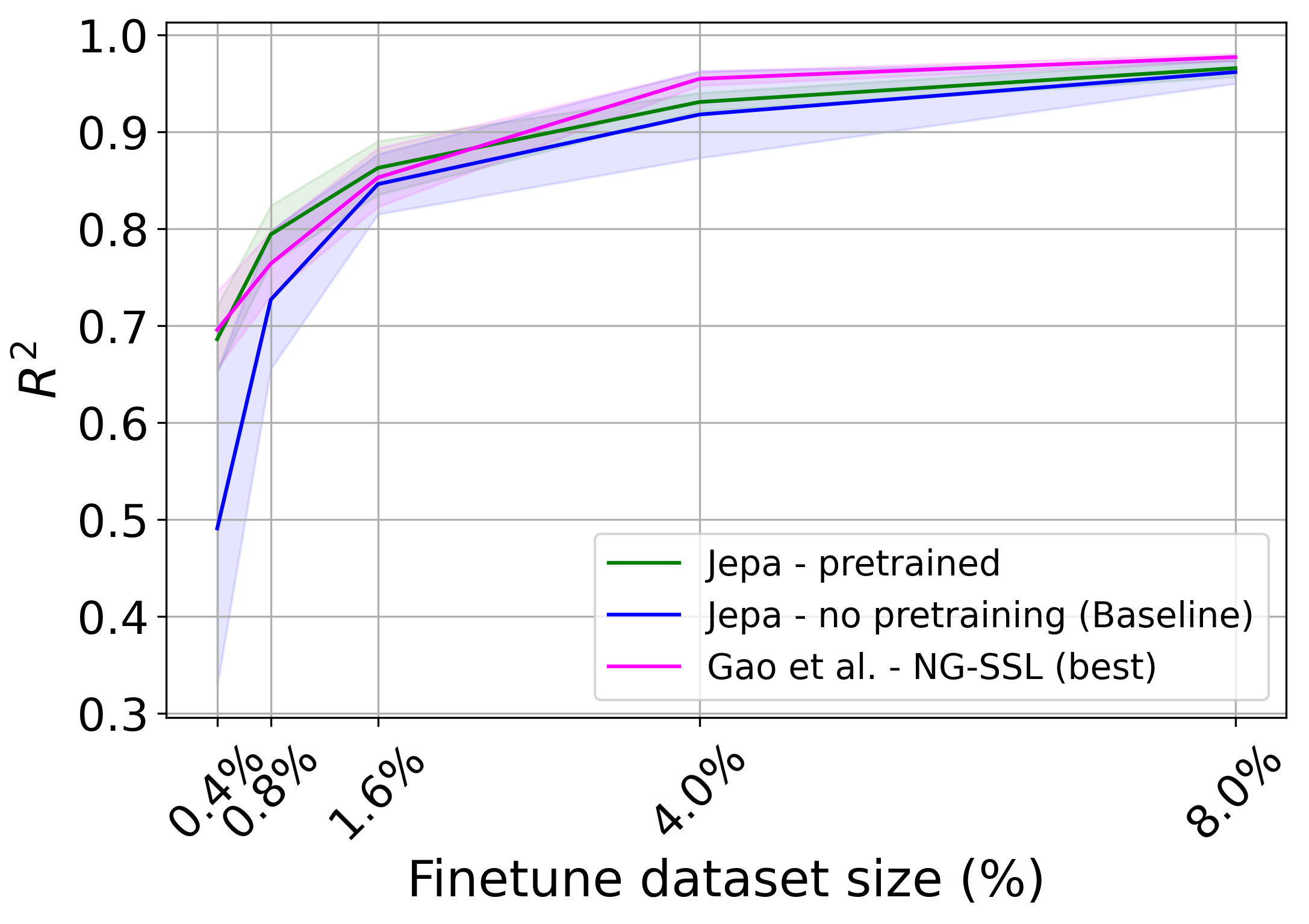}
    \caption{Comparison between our pretraining strategy and the best performing SSL model from \cite{gao2024self}. For different finetune dataset sizes we compare the $R^2$ for prediction of EA of the conjugated copolymer test data~\cite{aldeghiGraphRepresentationMolecular2022}.}
    \label{fig:comparison_JEPA_GaoBest_NoPretrain}
\end{figure}

Overall, the performance increase using input space self-supervised pretraining and embedding space self-supervised pretraining (JEPA, our method) is comparable. Figure~\ref{fig:comparison_JEPA_GaoBest_NoPretrain} reveals that our method is slightly better in the very low data scenarios, and Gao et al.'s method~\cite{gao2024self} performs better with more available labelled data. 

\begin{figure}[!htb]
    \centering
    \includegraphics[width=0.7\textwidth]{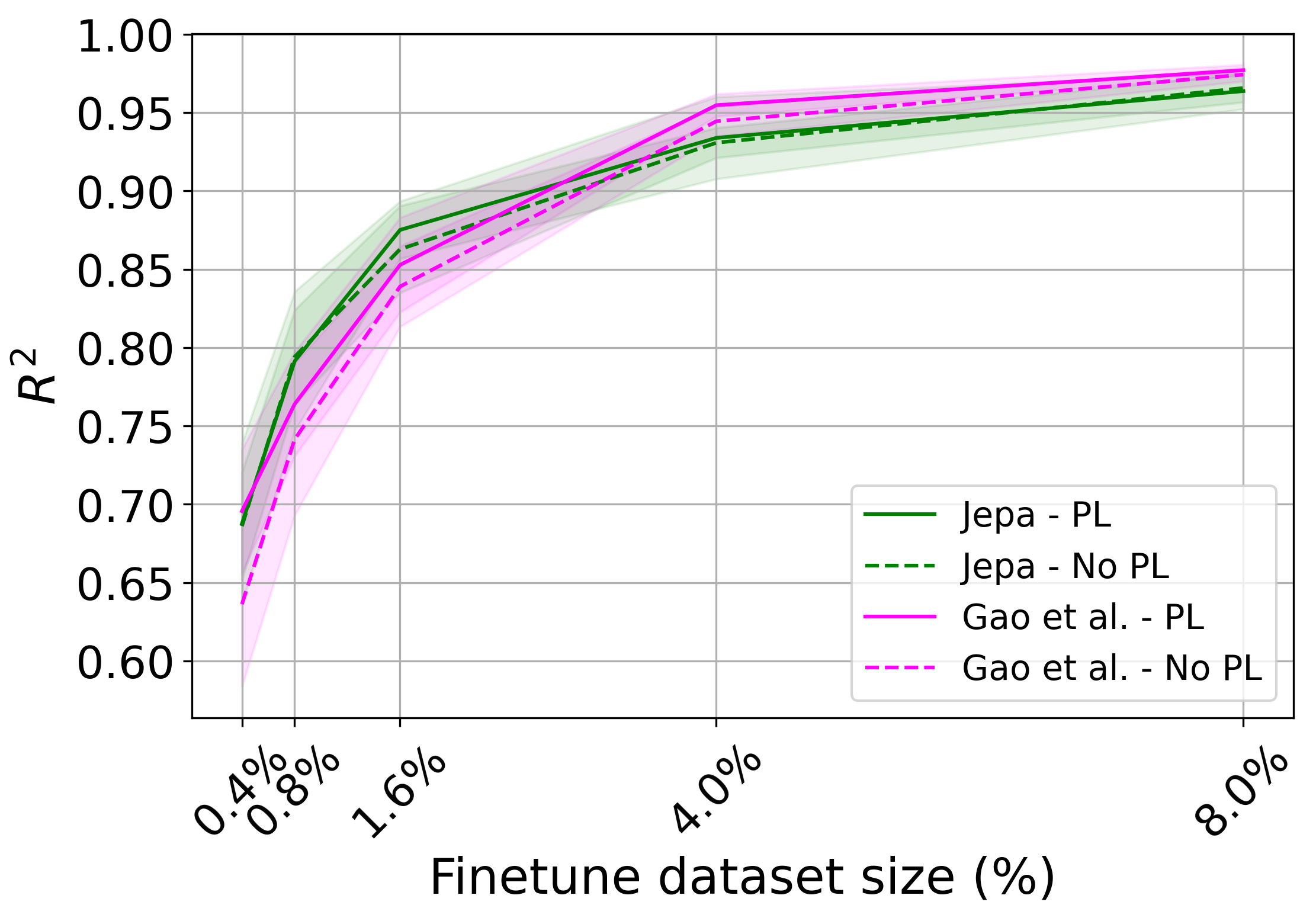}
    \caption{The effect of using an additional pseudolabel objective in input space SSL~\cite{gao2024self} and embedding space SSL (ours). For different finetune dataset sizes we compare the $R^2$ for prediction of EA of the conjugated copolymer test data~\cite{aldeghiGraphRepresentationMolecular2022}. We include both the scenarios when only the wD-MPNN encoder weights are transferred (No PL), and the scenario when also the pseudolabel (molecular weight) predictor weights are transferred (PL).}
    \label{fig:comparison_JEPA_GaoBest_PL}
\end{figure}

Using an additional pseudolabel objective as described in Section~\ref{sec:PL_method} leads to a consistent improvement in \cite{gao2024self}. While we observed significant improvements in single scenarios, overall the performance improvements attributed to using an additional pseudolabel objective are smaller for our method than for the input-space SSL approach as shown in Figure~\ref{fig:comparison_JEPA_GaoBest_PL}. This suggests that our strategy potentially already captures relevant information related to the polymer molecular weight pseudolabel. 

% In their work \cite{gao2024self}, they employ two SSL tasks: one at the node level, masking nodes and edges and learning to predict them, and the other at the graph level, predicting a pseudolabel corresponding to the molecular weight of the polymer, derived from the monomers' weights. They test both tasks separately and together, and they discover that pretraining via both tasks proves to be the most effective. This result aligns with findings in the literature \cite{hu2019strategies} that SSL on graphs works better when using both node-level and graph-level tasks together.
% OLD RESULTS: Their pretraining strategy seems more effective, yielding better performance. However, note that some experimental factors differed between our experiments and theirs, particularly in our use of different pretraining and finetuning datasets at every run, utilizing cross-validation. In contrast, the other study always uses the same datasets across runs.
% OLD RESULTS: Interestingly, both methods achieve a significant performance boost when also transferring the weights of the fully connected layers used to predict the molecular weight (the pseudolabel objective). This suggests that the knowledge obtained while learning to predict the molecular weight is particularly helpful for the task of predicting the electron affinity. 

\subsection{Comparison to random forest baseline}\label{sec:rf_comparison}
We also compare the performance of our model to a random forest model that uses fingerprints generated from polymer sequences as inputs. Typically, simpler ML models outperform complex ones in data-scarce scenarios. The wD-MPNN generates a polymer representation (fingerprint) through pooling node embeddings learned during training, which is then used as input for the MLP predictor. In contrast, the random forest model is trained on handcrafted fingerprints as inputs. We follow the approach in~\cite{aldeghiGraphRepresentationMolecular2022}, utilizing count-vector Extended-Connectivity Fingerprints (ECFP)  \cite{rogers2010extended} of size 2048 and radius 2, computed with RDKit \cite{landrum2006rdkit}. The polymer fingerprints are obtained by averaging across an ensemble of 32 oligomer sequence fingerprints. 

\begin{figure*}[!htb]
    \centering
    \begin{subfigure}[b]{0.48\textwidth}
        \centering
        \includegraphics[width=\textwidth]{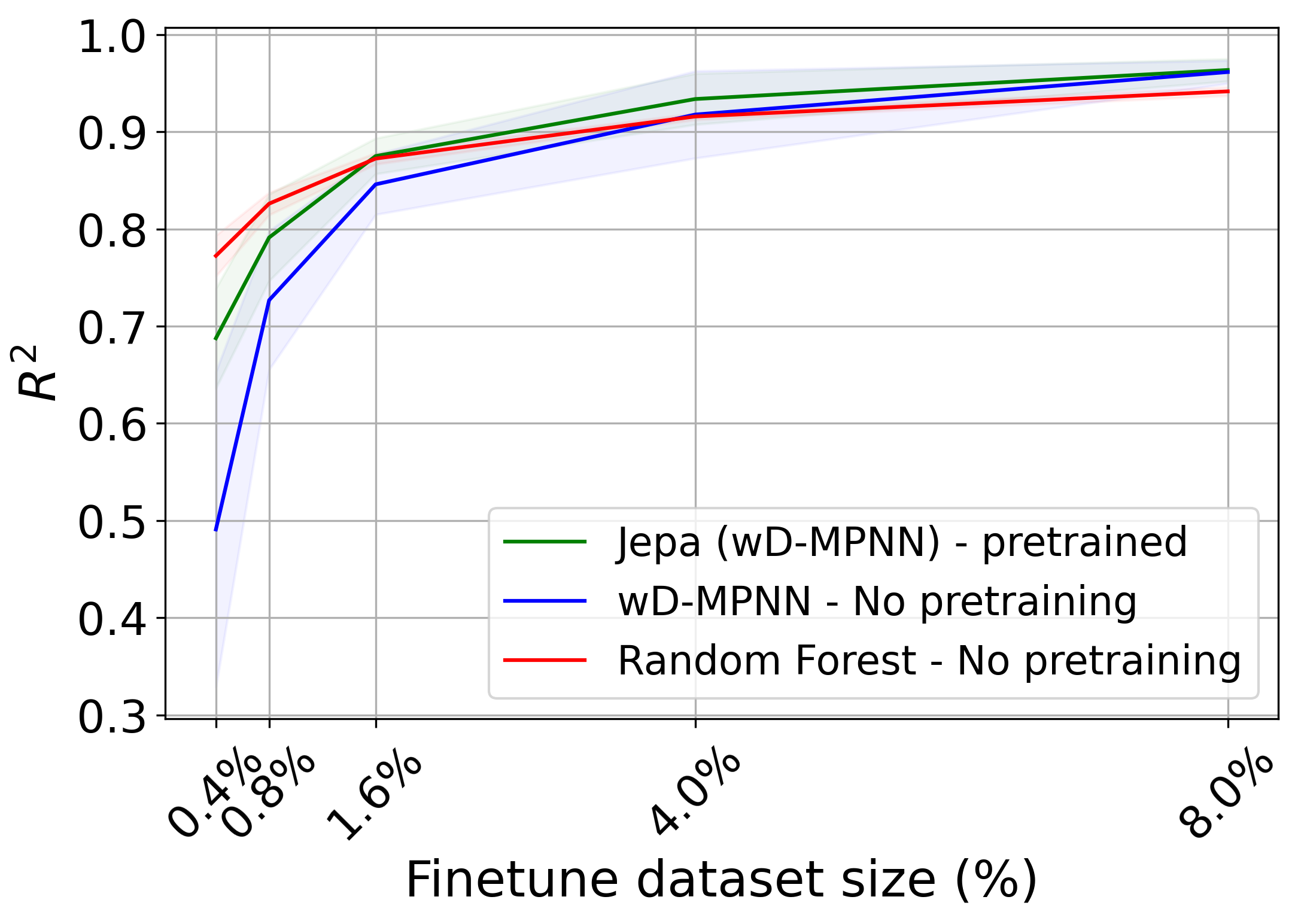}
    \caption{}
    \label{fig:RF_comparison_aldeghi}
    \end{subfigure} 
\hfill
    \begin{subfigure}[b]{0.48\textwidth}
        \centering
        \includegraphics[width=\textwidth]{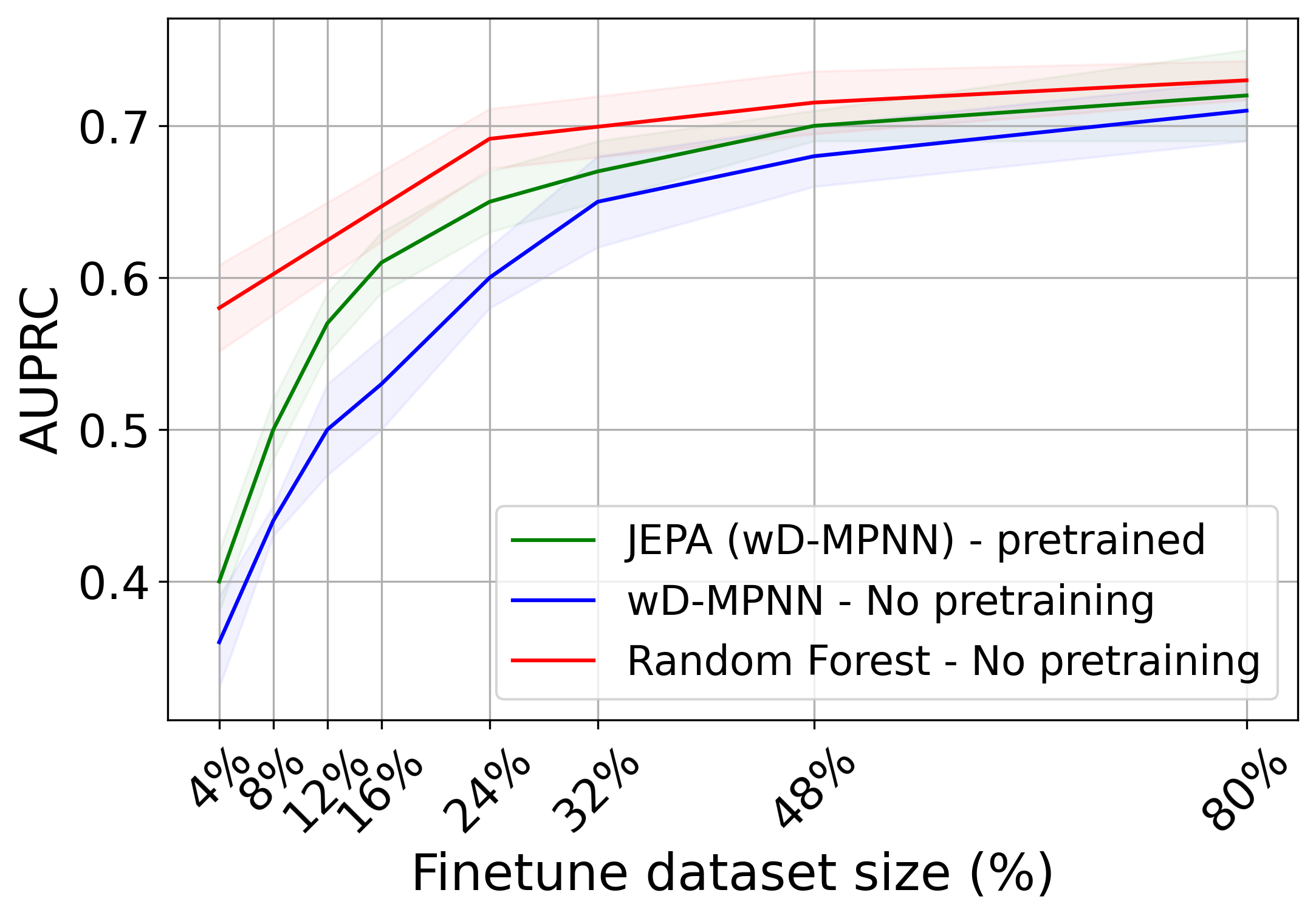}
    \caption{}
    \label{fig:RF_comparison_diblock}
\end{subfigure} 
\caption{(a) Comparison between our pretraining-finetuning strategy and a random forest model on the conjugated copolymer dataset~\cite{aldeghiGraphRepresentationMolecular2022}, predicting the EA property, for different finetune dataset sizes. (b) Comparison of phase behavior classification performance between our pretraining-finetuning strategy and a random forest model on the diblock copolymer dataset~\cite{arora2021dataset}, for different finetune dataset sizes.}
\end{figure*}

While the polymer-JEPA demonstrated improvements over our baseline model without pretraining in low-data regimes, we observed that the random forest model outperforms the pretrained wD-MPNN in the very low labeled data regimes. This trend is evident in Figure \ref{fig:RF_comparison_aldeghi}, where the random forest model exhibited an advantage over the pretrained wD-MPNN when using 0.4\% (192 points) and 0.8\% (384 points) of the dataset for finteuning. The advantage of the RF model vanishes as the dataset size increases. In practice, only for a small regime around 4.0\% finetune data, we observe that the pretrained model outperforms both the random forest and the wD-MPNN without pretraining.   \\
For the smaller Diblock dataset (Figure \ref{fig:RF_comparison_diblock}), we observed that the random forest model outperforms our model throughout all data scenarios. This may be due to the increased difficulty of the multi-label classification task. Further, as Aleghi and Coley~\cite{aldeghiGraphRepresentationMolecular2022} point out, simply the mole fraction, correlated with the volume fractions of the two blocks is highly informative for determining the copolymer phase. They trained a RF model on mole fractions only which outperformed the wD-MPNN (no data scarce scenarios) without providing information about the chemistry. However, this represents an atypical case: for most polymer property prediction tasks, such strong correlations with easily computed features are rare, and more expressive, chemistry-aware models are necessary. 

As a result, we advise to consider also the application of simpler models for small labeled datasets with relatively simple structure to property relationships. However, one key advantage of our proposed method lies in its ability to eliminate the reliance on handcrafted descriptors. By learning directly from the polymer graph structure, JEPA offers greater adaptability to new datasets without the need to tune fingerprints for specific polymer structures. Specifically, the used fingerprinting~\cite{aldeghiGraphRepresentationMolecular2022} which involves the generation of oligomer ensembles and averaging their fingerprint, requires thoughtful engineering by experts and computation time. 

Lastly, we hypothesize that more diverse pretraining datasets and more finetune data could further elevate the performance of our JEPA (wD-MPNN) pretrained model, beyond the two tasks covered in this study. 

\subsection{Subgraphing}\label{sec:results_subgraphing}
This section presents an ablation study on subgraphing hyperparameters, including context and target subgraph size, the number of targets, and the subgraphing algorithm. These results explore the sensitivity of each parameter by varying them individually while keeping others fixed. 

Each experiment involved pretraining the model on 40\% of the conjugated copolymer dataset (Section~\ref{sec:training_procedure})) and finetuning on 0.4\% (192 data points) to simulate a label-scarce scenario.
Overall, across ablation experiments, self-supervised pretraining with our model consistently improved downstream property prediction performance. For the ablation experiments related to context subgraph size, target subgraph size and number of targets, we used the random-walk algorithm for subgraph creation due to its direct control over subgraph size. 

Within the searched subgraphing settings, we observe the highest performance increase using random walk subgraphing, a context size of 60\% and one target with a subgraph size of 10\% of the polymer graph. 

\paragraph{Context Subgraph Size}
As shown in table~\ref{tab:context_size}, we identify an optimal context size of 60\% (relative to the full graph) for predictive tasks, balancing informativeness and overlap with the target. This corresponds to around 10-15 atoms of the full polymer graph (20 - 25 atoms). The model is robust to context size changes, with only minor performance drops at smaller sizes (e.g., 20\%). Based on our results, a context size of 50-75\% is suggested to be broadly effective across different size of molecular graphs.
In this experiment we use three target subgraphs, sized approximately 15\% of the total graph. 

\begin{table}[!htb]
  \centering
  \caption{Impact of context subgraph size on property prediction performance evaluated for the electron affinity in the conjugated copolymer dataset~\cite{aldeghiGraphRepresentationMolecular2022}.}
  \label{tab:context_size}
  \begin{tabular}{p{0.28\textwidth}p{0.28\textwidth}p{0.28\textwidth}}
    \toprule
    \textbf{Context size} & \textbf{$R^2\uparrow$} & \textbf{$RMSE$ $\downarrow$} \\
    \midrule
    \textit{No pretraining} & $\mathit{0.46 \pm 0.15}$ & $\mathit{0.44 \pm 0.06}$ \\
    20\% & $0.56 \pm 0.07$ & $0.39 \pm 0.03$ \\
    40\% & $0.60 \pm 0.06$ & $0.37 \pm 0.03$ \\
    \textbf{60\%} & $\mathbf{0.65 \pm 0.03}$ & $\mathbf{0.35 \pm 0.02}$ \\
    80\% & $0.62 \pm 0.07$ & $0.37 \pm 0.03$ \\
    95\% & $0.61 \pm 0.05$ & $0.37 \pm 0.02$ \\
    \bottomrule
  \end{tabular}
\end{table}

\paragraph{Target Subgraph Size}
Further, we find an optimal target subgraph size of 10\% of the total graph, which corresponds to two to three atoms in the considered polymer graphs (Table~\ref{tab:target_size}). We hypothesize that the ideal range ensures effective learning without oversimplifying (too small targets) or overcomplicating (too large targets) the prediction task. When applying this method to other polymer datasets with graphs of different sizes, the optimal range might differ slightly.
We hypothesize that a target size range of 10\% to 20\% should be effective across different molecular sizes. For smaller polymer graphs, the target size might lean towards 20\%, while for larger molecules, it might be closer to 10\%. This should avoid the extreme cases where the target is either too small (a single node) or too large (several molecular substructures together).

\begin{table}[!htb]
  \centering
  \caption{Impact of target subgraph size on property prediction performance evaluated for the electron affinity in the conjugated copolymer dataset~\cite{aldeghiGraphRepresentationMolecular2022}.}
  \label{tab:target_size}
    \begin{tabular}{p{0.28\textwidth}p{0.28\textwidth}p{0.28\textwidth}}
    \toprule
    \textbf{Target size} & \textbf{$R^2\uparrow$} & \textbf{$RMSE$ $\downarrow$}\\
    \midrule
    \textit{No pretraining} & $\mathit{0.46 \pm 0.15}$ & $\mathit{0.44 \pm 0.06}$ \\
    5\% & $0.61 \pm 0.07$ & $0.37 \pm 0.03$ \\
    \textbf{10\%} & $\mathbf{0.66 \pm 0.02}$ & $\mathbf{0.35 \pm 0.01}$ \\
    15\% & $0.65 \pm 0.03$ & $0.35 \pm 0.02$ \\
    20\% & $0.63 \pm 0.03$ & $0.36 \pm 0.02$ \\
    \bottomrule
  \end{tabular}
\end{table}

\paragraph{Number of target subgraphs}
The model is not highly sensitive to the number of targets (see Table~\ref{tab:n_of_targets}).
A single target provides the best performance, but we do not record a large drop if more targets are included.
Here, we use the optimal configuration of 60\% context size and 10\% target size as identified as suitable previously. We reason that fewer targets increase exposure to diverse subgraphs per epoch, improving generalization, while more targets risk overfitting by repeatedly predicting similar subgraphs.
\begin{table}[!htb]
  \centering
  \caption{Impact of the number of predicted target subgraphs on property prediction performance evaluated for the electron affinity in the conjugated copolymer dataset~\cite{aldeghiGraphRepresentationMolecular2022}.}
  \label{tab:n_of_targets}
  \begin{tabular}{p{0.28\textwidth}p{0.28\textwidth}p{0.28\textwidth}}
  \toprule
    \textbf{number of targets} & \textbf{$R^2\uparrow$} & \textbf{$RMSE$ $\downarrow$} \\
    \midrule
    \textit{No pretraining} & $\mathit{0.46 \pm 0.15}$ & $\mathit{0.44 \pm 0.06}$ \\
    \textbf{1} & $\mathbf{0.67 \pm 0.01}$ & $\mathbf{0.34 \pm 0.01}$ \\
    2 & $0.64 \pm 0.03$ & $0.36 \pm 0.01$ \\
    3 & $0.66 \pm 0.02$ & $0.35 \pm 0.01$ \\
    4 & $0.65 \pm 0.05$ & $0.35 \pm 0.02$ \\
    5 & $0.61 \pm 0.04$ & $0.37 \pm 0.02$ \\
    \bottomrule
  \end{tabular}
\end{table}

\paragraph{Subgraphing type}
We observe (Table \ref{tab:subgraphing_type}) that all algorithms perform similarly, with a slight advantage for the random-walk subgraphing, demonstrating the best performance and stability. Overall, we find that variations in subgraphs and subgraph sizes play a more crucial role in determining model performance than the chemical meaningfulness of the subgraphs themselves. 

Using a fixed context size of 60\% and target size of 10\% (single target), we compared the random-walk, motif-based, and METIS subgraphing.
Interestingly, the motif-based method, which leverages domain knowledge to produce chemically meaningful subgraphs, exhibits slightly lower performance than the other two algorithms. While motif-based subgraphing generates chemically meaningful subgraphs, it tends to produce a relatively small number of subgraphs, in a deterministic fashion, potentially limiting model generalization by increasing the likelihood of encountering similar or identical subgraphs (both context and target ones) throughout training. On the other hand, the METIS algorithm, while also producing consistent subgraphs at each epoch, generates on average a higher number of subgraphs compared to the motif-based approach, introducing more variability across epochs.
Finally, random-walk subgraphing generates different subgraphs at every epoch, thanks to the stochastic nature of the subgraphing process. Moreover, it allows for greater control over subgraph sizes, a factor previously highlighted as relatively influential in model performance.

\begin{table}[!htb]
  \centering
  \caption{Impact of different subgraphing algorithms on property prediction performance evaluated for the electron affinity in the conjugated copolymer dataset~\cite{aldeghiGraphRepresentationMolecular2022}.}
  \label{tab:subgraphing_type}
  \begin{tabular}{p{0.28\textwidth}p{0.28\textwidth}p{0.28\textwidth}}
    \toprule
    \textbf{Subgraphing} & \textbf{$R^2\uparrow$} & \textbf{$RMSE$ $\downarrow$} \\
    \midrule
    \textit{No pretraining} & $\mathit{0.46 \pm 0.15}$ & $\mathit{0.44 \pm 0.06}$ \\
    Motif-based & $0.63 \pm 0.05$ & $0.36 \pm 0.02$ \\
    Metis & $0.67 \pm 0.04$ & $0.34 \pm 0.02$ \\
    \textbf{Random Walk (RW)} & $\mathbf{0.67 \pm 0.01}$ & $\mathbf{0.34 \pm 0.01}$ \\
    \bottomrule
  \end{tabular}
\end{table}

\section{Conclusion}
This study introduces a novel self-supervised pretraining strategy for the polymer domain, where labeled data is often scarce. The presented method operates on polymer molecular graphs, leveraging the concept of JEPAs. Our study stands as one of the initial efforts in exploring JEPAs for molecular graph-related tasks, thus enriching the understanding and analysis of this new architectural family in the molecular domain. We provide guidelines for subgraphing; in particular on size and algorithm selection.
Our experiments show that self-supervised pretraining on conjugated copolymer data consistently improves the downstream prediction accuracy for a dataset that describes a different polymer space (diblock copolymers), showcasing the ability of transferring general knowledge across polymer datasets of different applications.
When pretraining and finetuning on the same polymer space of conjugated copolymers our method helps in label-scarce data scenarios up to up to around 8\% (ca. 3440 polymers) of labeled data availability. The performance improvement (in R2) varies from 39.8\% in the smallest labeled  data scenario to 0.4\% in the scenario with 8\% labelled data. \\
Comparing our polymer-JEPA self-supervised model (embedding space) with node/edge-masking self-supervised learning (input space), we observe that we achieve comparable performance on the tested downstream task. Further, both methods benefit from including a pseudotask prediction (molecular weight), with the benefit being less pronounced in our embedding space self-supervised pretraining strategy. Additionally, we showed that a simple baseline random forest model can outperform our method in certain scenarios in the two downstream tasks considered. However, this method relies on expert-engineered polymer fingerprints, which is a not necessary with our method that learns from the polymer graph directly. \\
Looking ahead, the embedding-based nature of JEPA offers promising opportunities for integrating multimodal data and utilizing a variety of different experimental and synthetic datasets for pretraining. Generally, we hypothesize that more diverse pretraining datasets could contribute to further increasing the performance of our JEPA (wD-MPNN) model compared to not using a pretrained model, also beyond the two tasks covered in this study. 

%Old conclusion
%Through our experiments, we demonstrated the efficacy of our pretraining approach in significantly enhancing performance, particularly in scenarios where finetune data is very limited. However, we also showed how other SSL approaches \cite{gao2024self} or a simpler model (Appendix \ref{RF}) outperform our method, in the two downstream tasks considered. Interestingly, our pretraining strategy exhibits generalization capabilities, as evidenced by its effectiveness on a dataset representing a different chemical space from the one utilized for pretraining. Hence, the proposed method could prove useful in complex classification or regression tasks, where labeled data is scarce, and pretraining with the proposed method could lead to a positive knowledge transfer.

%%%%%%%%%%%%%%%%%%%%%%%%%%%%%%%%%%%%%%%%%%%%%%%%%%%%%%%%%%%%

%\bmhead{Supplementary information}

%If your article has accompanying supplementary file/s please state so here. 

%\bmhead{Acknowledgements}

%Acknowledgements are not compulsory. Where included they should be brief. Grant or contribution numbers may be acknowledged.

\section*{Declarations}

%For Journal of Cheminformatics: 

%\begin{itemize}
%\item Availability of data and materials
%\item Competing interests
%\item Authors' contributions
%\item Acknowledgements
%\end{itemize}

\paragraph{Availability of data and materials}
The used data and code are provided in the Github repository \url{https://github.com/Intelligent-molecular-systems/Polymer-JEPA}. 
\paragraph{Competing interests}
Not applicable
\paragraph{Funding}
Not applicable
\paragraph{Authors' contributions}
FP: Conceptualization, Methodology, Software (lead), Formal analysis, Writing – original draft.
GV: Supervision, Methodology, Software, Formal analysis, Writing – original draft.
JW: Conceptualization, Supervision, Writing – review and editing.
\paragraph{Acknowledgements} 
We thank Qinghe Gao for the discussions about SSL on polymer graphs.  

\bigskip

\begin{appendices}

\section{Additional subgraphing requirements}
\label{app:subgraph_reqs}
The subgraphs used as inputs to our model comply with the following requirements: 
 \begin{enumerate}
    \item  In the case of a copolymer, the context subgraph should include elements from both monomers: predicting a part of monomer B, if monomer B is missing from the context is not possible, 
\item  Every edge and every node should be in at least one subgraph \cite{GraphViTMLPMixer2023, skenderiGraphlevelRepresentationLearning2023} to include full global information and full input representation,
\item  The context patch (subgraph) should be larger, hence more informative, than the targets patches (subgraphs) we are trying to predict from the context \cite{assranSelfSupervisedLearningImages2023},
\item  The target subgraphs should have minimal overlap with the context subgraph \cite{assranSelfSupervisedLearningImages2023, skenderiGraphlevelRepresentationLearning2023} to make the prediction task less trivial,
\item  The context and targets subgraphs should change at every training loop to prevent overfitting \cite{assranSelfSupervisedLearningImages2023, skenderiGraphlevelRepresentationLearning2023},
\item  For every directed edge $e_{vu}$ in a subgraph, include also the edge $e_{uv}$, to comply with encoder architecture (wD-MPNN~\cite{aldeghiGraphRepresentationMolecular2022}).
\end{enumerate}

%\begin{figure}[htbp]
%    \centering
%    \includegraphics[width=\textwidth]%{Images/polymer_JEPA.png}
%    \caption{Simplified model representation. }
%    \label{fig:polymer_JEPA}
%\end{figure}

\section{Node-centred and Edge-centred message passing with wD-MPNN}\label{app:node_vs_edge_WDMPNN}
The choice of edge-centred convolutions differs from most common GNNs, which use node-centred message passing. The motivation behind this design, explained in \cite{dmpnn2019} is to prevent totters, that is, to avoid messages being passed along any path of the form $v_1v_2...v_n$ where $v_i = v_{i+2}$ for some $i$, which are thought to introduce noise into the graph representation by creating unnecessary loops in the message passing trajectory. However, we hypothesize that such loops are not less relevant for the dataset considered, as the polymer graphs utilized are unlikely to contain such small cyclic patterns. Working with edge-centred convolutions requires big and sparse adjacency matrices of shape $A_{e \times e}$ and $A_{e \times n}$, which makes the training process more cumbersome, also due to the edge-centred convolution not being well supported in \textit{Pytorch Geometric}. 
We compare the performance of the convolution methods predicting the EA and IP of the conjugated copolymer dataset in Tables~\ref{tab:node_vs_edge_EA} and~\ref{tab:node_vs_edge_IP} respectively. The results were obtained in the scenario where we train end-to-end on 80\% of the data, and test on 20\%. As visible in the tables, both methods achieve the same results, making the node-centred convolution a good solution, given the ease of implementation.
\begin{table}[!htb]
  \centering
    \caption{Comparing the performance of the node-centred and the edge-centred message passing model on prediction of the electron affinity.}
  \label{tab:node_vs_edge_EA}
  \begin{tabular}{lcc}
    \toprule
    \textbf{wD-MPNN} & \textbf{R2 $\uparrow$} & \textbf{RMSE $\downarrow$} \\
    \midrule
    Edge-centred & $0.998 \pm 0.0003$ & $0.029 \pm 0.002$ \\
    Node-centred & $ 0.998 \pm 0.0002$ & $0.027 \pm 0.001$ \\
    \bottomrule
  \end{tabular}
\end{table}
\begin{table}[!htb]
  \centering
    \caption{Comparing the performance of the node-centred and the edge-centred message passing model on prediction of the ionization potential.}
  \label{tab:node_vs_edge_IP}
  \begin{tabular}{lcc}
    \toprule
    \textbf{wD-MPNN} & \textbf{R2 $\uparrow$} & \textbf{RMSE $\downarrow$} \\
    \midrule
    Edge-centred & $0.998 \pm 0.0007$ & $0.022 \pm 0.004$ \\
    Node-centred & $ 0.997 \pm 0.0004$ & $0.025 \pm 0.003$ \\
    \bottomrule
  \end{tabular}
\end{table}
\newpage

\end{appendices}


\bibliographystyle{unsrtnat}
\bibliography{references}% common bib file

\end{document}